# 2D View Aggregation for Lymph Node Detection Using a Shallow Hierarchy of Linear Classifiers

Ari Seff, Le Lu, Kevin M. Cherry, Holger Roth, Jiamin Liu, Shijun Wang, Joanne Hoffman, Evrim B. Turkbey, and Ronald M. Summers

Imaging Biomarkers and Computer-Aided Diagnosis Laboratory,
Radiology and Imaging Sciences, National Institutes of Health Clinical Center,
Bethesda, MD 20892

**Abstract.** Enlarged lymph nodes (LNs) can provide important information for cancer diagnosis, staging, and measuring treatment reactions, making automated detection a highly sought goal. In this paper, we propose a new algorithm representation of decomposing the LN detection problem into a set of 2D object detection subtasks on sampled CT slices, largely alleviating the curse of dimensionality issue. Our 2D detection can be effectively formulated as linear classification on a single image feature type of Histogram of Oriented Gradients (HOG), covering a moderate field-of-view of 45 by 45 voxels. We exploit both simple pooling and sparse linear fusion schemes to aggregate these 2D detection scores for the final 3D LN detection. In this manner, detection is more tractable and does not need to perform perfectly at instance level (as weak hypotheses) since our aggregation process will robustly harness collective information for LN detection. Two datasets (90 patients with 389 mediastinal LNs and 86 patients with 595 abdominal LNs) are used for validation. Cross-validation demonstrates 78.0% sensitivity at 6 false positives/volume (FP/vol.) (86.1% at 10 FP/vol.) and 73.1% sensitivity at 6 FP/vol. (87.2% at 10 FP/vol.), for the mediastinal and abdominal datasets respectively. Our results compare favorably to previous state-of-the-art methods.

## 1   Introduction.

Lymph nodes (LNs) play a crucial role in disease progression and treatment. Enlarged lymph nodes in particular, considered by the widely followed RECIST criteria to be at least 10 mm in short axis diameter [1], are considered suspicious and can indicate metastatic cancer. Radiologists routinely assess lymph nodes in the vicinity of tumors to monitor patient response to various therapies. As a manual task, this can be highly time consuming and error prone. Thus, there have been intensive studies on automatic detection of lymph nodes on CT images in different sections of the body.

Previous work mostly leverages the direct 3D information from volumetric CT images. For instance, [2, 3] exploit the mixture of 3D Hessian blobness filter, directional difference filter, shape morphology and volume thresholds. The state-of-the-art methods [4, 5] perform boosting-based feature selection and integration over a pool of 50~60 thousands of 3D Haar wavelet features to finally obtain a strong binary classifier on selected features. Due to the limited available training data and the intrinsic high dimensionality of modeling on complex 3D CT features, 3D LN detection is non-trivial.

Particularly, lymph nodes have large within-class appearance/location/pose variations and low contrast from surrounding anatomy over a patient population. This results in many false positives to assure moderately high detection sensitivity [3, 6] or only limited sensitivity levels [5, 7]. The good sensitivities achieved at low FP range in [4] are not comparable with the other studies since [4] reports on axillary and pelvic + abdomen body areas, and others evaluate on either mediastinum [2, 5, 6] or abdomen [3, 7].

The essential idea of this work, LN detection by aggregating 2D views, assumes at least some portion of the 2D image patterns (on orthogonal slices) can be encoded and detected reliably for any true lymph node residing in a 3D volume of interest (VOI), while no or very weak 2D detections may be found for a false LN subvolume. The 2D view-based LN detection problem may contain labeling noise (as the label is given per VOI) but inhabits a lower dimensional feature space, with one order of magnitude more samples for training, compared with 3D detection. Our 2D detector is effectively implemented (following a 3D candidate generation preprocessing step) using Liblinear [8] on a single image feature type of Histogram of Oriented Gradients [9, 10]. We exploit simple pooling and sparse linear weighting schemes (**Sec. 2.3**) to softly aggregate these 2D detection scores for the final 3D LN detection. Importantly, we do not need to classify all 2D slices from a 3D lymph node VOI correctly or with an ultra high accuracy to obtain good results on LN detection. However any single detection error of 3D VOIs [4, 5] causes either a missing lymph node or a false positive count per case.

Our main contributions are three-fold. First, we present a new lymph node detection approach in 3D CT images by running a 2D detector on orthogonal slice views and aggregating their scores per VOI to compute the final LN classification confidence. Second, instead of deep cascade boosting classifiers [4, 5], our 2D detector works as a single shallow template matching step through the efficient inner-product between classifier and image in HOG feature space. Third, *to the best of our knowledge, we are the first to formulate the 3D lymph node classification problem as a sparse linear fusion of detections running only on 2D CT views*. Unlike [4, 5], our method does not need explicit segmentation for lymph node detection. Our method reports good performance on two datasets (90 patients with 389 mediastinal LNs and 86 patients with 595 abdominal LNs), and compares favorably to prior state-of-the-art work in mediastinal [2, 5, 6] and abdominal [3, 7] LN detection. The proposed method is suitable for detecting small, scattered anatomical objects in 3D scans, including lymph nodes.

## 2      Methods

### 2.1      Candidate Generation (CG) as Preprocessing

The first phase of the lymph node detection system involves the generation of a list of volumes of interest, containing all enlarged LNs as targeted objects (at the expense of low specificity), from any input 3D CT image. Within the body search region, four primitive types of voxel-level features are calculated at down-sampled grid space (every 3rd voxel in $(x, y, z)$): intensity, multiscale Hessian blobness scores, response values from multiscale DOG (Difference of Gaussian) filters, and the averages of these feature values from the neighborhoods of 3, 6, and 12 voxels as radii. In this way, multiscale low level image features are densely computed on the 3x3x3 grid voxels in CT volumes and used to further train a random forest [11] classifier, based on the manually segmented LN masks for classifying positive or negative class voxels (i.e., voxels inside an LN mask are treated as positive, and vice versa). Thus, a probability map is generated by the random forest (RF) for each CT scan which is thresholded and spatially grouped to obtain a set of detection candidates. The candidate location is recorded as the centroid of the grouped voxels. Each candidate is cropped as a cube VOI of $45 \times 45 \times 45$ voxels, centered at its found location and then assigned the label. If its location is inside a ground truth LN mask, the corresponding candidate is labeled as +1, otherwise -1. Through this step, close to 100% LN sensitivity can be achieved at 35~40 FPs per case by setting a moderately conservative threshold calibrated from the training RF Receiver operating characteristic (ROC) curve. Given sufficient training voxels and enough trees for the RF (e.g., 50~200), such a performance goal is feasible and may be possible through other ways of preprocessing, which is not the core topic in this paper. *Note that [5] boosts complex 3D HAAR wavelet features to form a one-shot LN detection system which has better sensitivity at low FP range, but their maximal sensitivity saturates at 65%. We use more primitive 3D Hessian/DOG features under a less greedy classifier to assure very high sensitivity only at high FP rates.*

### 2.2 3D Detection Decomposition as a Set of 2D Detections

**View Sampling:** From above, each candidate $V$ has a computed centroid location $(x, y, z)$ in 3D CT coordinates. From the center of $V$, for simplicity, we take 2D slices or views at $45 \times 45$ voxels along each of the three coordinate axes (i.e., axial, coronal and sagittal slices). After evenly sampling at 0, 1, 2, 3, ..., and $k$ voxels away from the centroid we have 27 total image views $\{v_i\}_{i=1,2,...27}$ per candidate (without loss of generality, we set $k = 4$): stacking 9 sagittal, coronal, and axial slices from along x, y, and z-axes respectively. We also transfer the +1/-1 label from $V$ to $\{v_i\}_{i=1,2,...27} \in V$ and attempt to build an effective detector on 2D views of $\{v_i\}_{i=1,2,...27}$ for all $V$, obtained from CG preprocessing. For generality, our detector will be learned by treating each $v_i$ as an independent instance, regardless of its VOI and patient affiliations. **Fig. 1** demonstrates an example of view sampling from a mediastinal lymph node candidate.

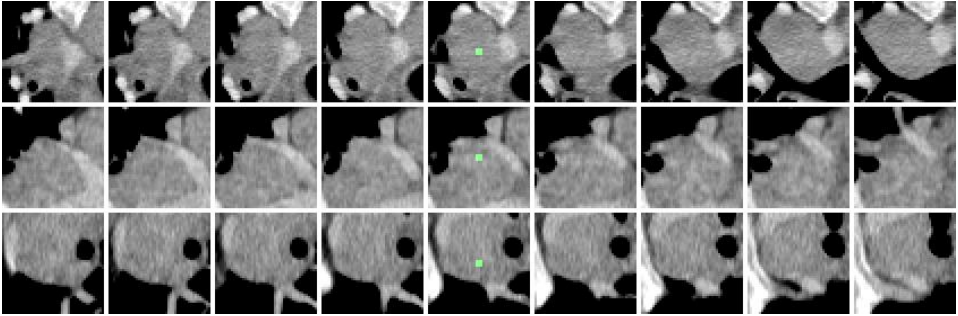

**Fig. 1.** Example mediastinal lymph node candidate with 9 consecutive axial (top row), coronal (middle row), and sagittal slices (bottom row). The candidate centroid is shown in green in the center column.

**Feature Extraction:** Detecting lymph node appearance against surrounding context in CT images is normally addressed by calculating 3D contrast filters such as 3D minimum directional difference filters [2, 3] or Haar features [4, 5]. In certain 2D views or slices, the intensity contrast pattern inside and outside of a lymph node can be effectively captured on the gradient domain as well, via multi-resolution Histogram of Oriented Gradients (HOG) features [9, 12], as shown in **Fig. 2**. HOG features divide an image window to be encoded into square cells, delineating the quantized magnitude and orientation distributions of local intensity gradients for each cell. There are 13 HOG features after Principal Component Analysis-based compression, augmented with contrast sensitive and contrast insensitive features, leading to a 31-dimensional feature vector [9] per cell. Our multi-resolution HOG descriptor covers a moderately large 2D window of $45 \times 45$ voxels per view/slice, containing most lymph nodes of various sizes along with sufficient spatial context. The window can be configured with different cell sizes and numbers. For example, our implementation can extract $5 \times 5$ cells and 31 features per cell resulting in 775 features per image region, or $9 \times 9$ cells with 2511 features, mapping $v_i$ into HOG feature space $x_i \in \mathcal{R}^d$. Illustrative examples of a CT slice and its HOG feature maps in different cell configurations are shown in **Fig. 2**.

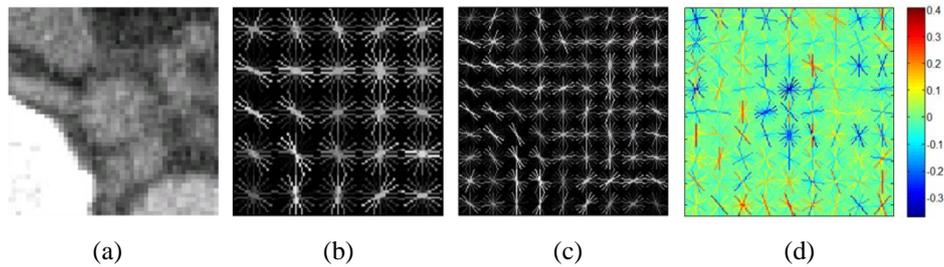

(a)          (b)          (c)          (d)

**Fig. 2.** Abdominal lymph node axial slice (a) and visual renderings of corresponding HOG features with $5 \times 5$ cells (b) and $9 \times 9$ cells (c) using VLFeat [10]. The learned feature weight vector $\omega$ in $\mathcal{C}_{9 \times 9}^1$ is also visualized in (d) for $9 \times 9$ cell HOG. The negative (blue) weights in the center of the abdominal LN model indicate expected low-magnitude intensity gradients.

**Efficient Linear Classification:** HOG features are normally coupled together with linear or non-linear (e.g., radial basis function (RBF) kernel) Support Vector Machine (SVM) classifiers for object detection [9, 12]. Taking our mediastinum dataset as an example, we have 4,168 VOIs from 90 patients after CG. By sampling 27 views per $V$, there are 112,536 2D view instances $\{x_i\}_{i=1,2,\dots,112536}$ for classification training and testing. However, 2D slice labels may be ambiguous and contain noise, requiring a robust classifier for effective handling as we simply label all slices from a TP-VOI as positive and vice versa. Some 2D views can be challenging to classify solely based on the local appearance, especially considering the CG process may not locate the true LN centroid.

For good efficiency and generality we enforce on linear classifiers, trained using Liblinear [8] which can effectively address the large-scale, robust linear classifier training issue. 2D view HOG feature vectors $\{x_i\}$ are treated as separate instances, looking to assign an individual confidence score to each. Given $l$ training instances $x_i \in \mathcal{R}^d, i = 1, \ldots, l$, and their corresponding $y_i \in \{-1, +1\}$ class labels, the L2-regularized and L2-loss linear SVM from Liblinear, $\mathcal{C}^1$, requires the minimization of the following cost:

$$\min_\omega \frac{1}{2}\omega^T\omega + C \sum_{i=1}^{l}(max(0, 1 - y_i\omega^T x_i))^2 \qquad (1)$$

The weight vector $\omega$ is then used to assign confidence scores to each instance in testing as $\omega^T x_i$, and its sign indicates the classification label. We further convert the confidence to a pseudo-likelihood probability $\in [0,1]$ by Sigmoid transform (**Eq. 2**), to be used next for view classification score aggregation.

$$p_i = Sigmoid(\omega^T x_i) = \frac{1}{1+\exp(-\omega^T x_i)} \qquad (2)$$

Liblinear has shown to be very robust with respect to a range of $C$ [8]. Our experimental results reported in this paper are based on $C = 1$. The feature weight vector $\omega$ learned for $9 \times 9$ cell HOG is visualized in **Fig. 2-d.** For comparison, a nonlinear RBF kernel SVM classifier, following a grid search for optimal parameters $C$ and kernel width $\sigma$, is also trained [13]. It performs slightly better than our Liblinear model in training, but degenerates greatly in validation indicating poor generality.

### 2.3 Detection Aggregation by Simple Pooling or Sparse Linear Fusion

After **Sec. 2.2**, there are 27 scores $\{p_i\}$ per $V$. In the evaluation of various sparse coding models for object recognition, max-pooling shows good performance, analogous to the V1 area of the mammalian visual cortex [14]. In our two-layer, shallow classification hierarchy, the maximum of the 27 confidence scores or probabilities can be reassigned to the candidate $V$ as its probability of being a lymph node.

$$\rho(V) = max_{i=1,2,\ldots 27}\{p_i\} \qquad (3)$$

Additionally, reassigning the arithmetic mean of the view-level confidence scores to each candidate (mean-pooling) is also tested in **Sec. 3.**

Treating max-pooling and mean-pooling as special cases, we propose to fit a sparse linear weighting function to the vector $P = [p_1, p_2, \ldots, p_{27}]^T$ and $\rho(P_k) = Sigmoid(\mathcal{W}^T P_k)$ where $\mathcal{W}$ is optimized according to **Eq. 4** with a Gaussian prior $G(\mathcal{W}|0, \Sigma)$ and k=1,2,...,M is the index of VOIs. By mapping $y_k = -1$ to $y_k = 0$ for VOI labels,

$$\mathcal{W} = \text{argmax}_\mathcal{W} \left[ \left[\sum_{k=1}^{M} y_k \log(\rho(P_k)) + (1 - y_k)\log(1 - \rho(P_k))\right] - \frac{w^T \Sigma^{-1} w}{2} \right] \qquad (4)$$

The hyper-parameters in $\Sigma$ as a diagonal matrix control the variance of individual elements in $\mathcal{W}$. When the $j$th diagonal coefficient $\sigma_j \to 0$ in $\Sigma$, the corresponding $\mathcal{W}_j = 0$ due to the zero variance, and $p_i$ becomes irrelevant for the final detection probability $\rho(P)$. This is known as the type-II maximum likelihood method in Bayesian statistics where $\Sigma$ and $\mathcal{W}$ can be effectively solved by two-loop iterative optimization [15] to obtain the linear classifier $\mathcal{C}^2$. In our shallow hierarchy, $\mathcal{C}^2$ is trained using the outputs from view level $C^1$. Max and mean-pooling are invariant to the view ordering in $P$ from $C^1$. We also sort $P$ ascendingly to align $C^1$ scores before $C^2$ training and testing.

In 6-fold CV (**Sec. 3**), the number of surviving non-zero coefficients in $\mathcal{W}$ varies $\in \{3, 4, \ldots, 8\}$ out of a total 27 dimensions which results in a sparse linear model. The reason for imposing the sparseness constraint on $\mathcal{W}$ is that elements of $P = [p_1, p_2, \ldots, p_{27}]^T$ are highly inter-dependent since $\{v_i\}$ are sampled slice by slice.

## 3  Experiments

**Data:** We collect two datasets[1] for mediastinum and abdomen lymph node detection (summarized in **Table 1**). The population for study is selected from patients scanned within a four-month period in 2012, showing lymphadenopathy in either target region. A lymph node is defined as enlarged if its short axis diameter is $\geq 10mm$ [1]. CT slice thickness varies from 1 mm to 1.25 mm, and axial in-plane image resolution varies from 0.63 mm to 0.97 mm. The use of the data is IRB approved.

Table 1. Lymph node detection datasets.

| *LN dataset* | *#Patients* | *#LNs* | *#TP Candidates* | *#FP Candidates* |
|---|---|---|---|---|
| Mediastinal | 90 | 389 | 960 | 3,208 |
| Abdominal | 86 | 595 | 1,005 | 3,484 |

**Protocol:** Six-fold cross validation (CV) is carried out by splitting the mediastinum and abdomen LN datasets separately into six disjoint sets at the patient level. Candidate generation (**Sec. 2.1**), trained previously, is not counted for this evaluation. Training classifiers $C^1$ and $C^2$ on 5 sets for a single CV iteration takes about 5 minutes. Processing time following candidate generation on a new testing patient case is generally $1\sim 3$ seconds (with HOG feature computation).

**Slice-level $C^1$ Performance:** At the slice level, 6,030 out of 25,920 positive class slice instances in the mediastinal dataset are classified correctly if taking $p_i = 0.5$ as a preliminary cutoff ($AUC = 0.719$). This results in a mean of 6.3 positively classified slices per positive VOI, in contrast to 1.5 slices per negative VOI. We perform the Kolmogorov-Smirnov test on the $C^1$ values between the positive and negative samples in validation. The obtained p-value is $< 0.01$, indicating a statistically significant difference. Thus, despite a relatively low recall (at slice-level), this layer of the classifer, $C^1$, can weakly differentiate between positive and negative 2D views, paving the way for the next step, $C^2$, to exploit slice score aggregation for VOI-level classification. In this layer, we implement varying spatial configurations of classifiers including $C^1_{3\times 3}$, $C^1_{5\times 5}$, and $C^1_{9\times 9}$ (illustrated in **Fig. 2**) with the best results reported using $C^1_{5\times 5}$ (see **Sec. 3: Optimal HOG Resolution** for a comparison of these configurations.)

**VOI-level $C^2$ Performance & Comparison:** As shown in **Fig. 3**, we report six-fold cross-validation (CV) FROC curves for both mediastinal and abdominal LN detection datasets. On validation, 63.1% sensitivity at 3 false positives/volume (FP/vol), 78.0% at

---

[1] Datasets will be made publicly available at http://clinicalcenter.nih.gov/drd/summers.html.

2D View Aggregation for Lymph Node Detection

6 FP/vol, and 86.1% at 10 FP/vol are achieved for the mediastinal datasets. These correspond to 57.8% sensitivity at 3 FP/vol, 73.1% at 6 FP/vol and 87.2% at 10 FP/vol, for the abdominal datasets. Numerical comparison of our method to previous work [3, 4, 5, 6, 7] is given in detail in **Table 2**. Our results are demonstrated to have 10%∼17% higher sensitivities (at 3, 6 FP/vol) than the recent state-of-the-art method in mediastinum [5], and ∼21% higher (at 13 FP/vol)) than the most recent work [7] in abdomen. Note that the results in [4] are not directly comparable to the rest due to different target body regions. Sparse linear fusion by $\mathcal{C}^2$ dominates over mean and max-pooling, which itself outperforms previous work, in the full range of the FROC curves.

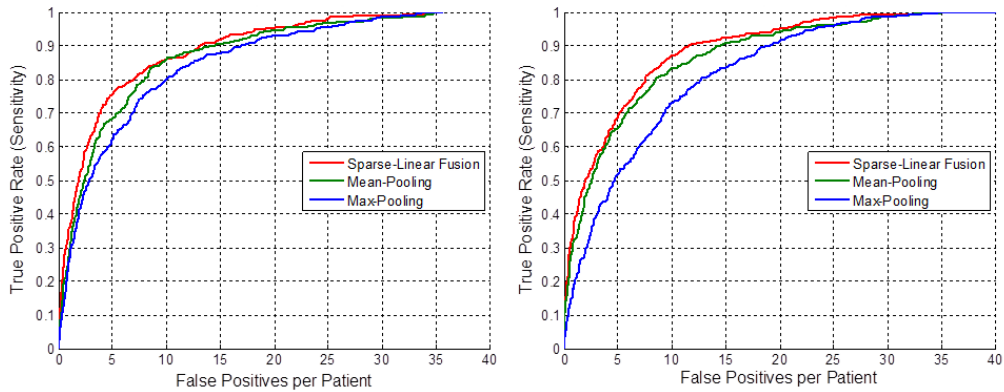

**Fig. 3.** Six-fold cross-validation FROC curves for the mediastinal (left) and abdominal (right) LN detection.

**Table 2.** Comparison of our method with other previous work on lymph node detection.

| *Method* | *Target Area* | *#Vol.* | *#LN* | *#TP* | *TPR(%)* | *FP/ vol.* |
|---|---|---|---|---|---|---|
| Kitasaka[3] | Abdomen | 5 | 221 | 126 | 57.0 | 58 |
| Barbu [4] | Pelvic + Abdomen | 54 | 569 | 455 | 80.0 | 3.2 |
| Feuerstein[6] | Mediastinum | 5 | 106 | 87 | 82.1 | 113 |
| Feulner [5] | Mediastinum | 54 | 289 | 153 | 52.9 | 3.1 |
| Feulner [5] | Mediastinum | 54 | 289 | 176 | 60.9 | 6.1 |
| Nakamura [7] | Abdomen | 28 | 95 | 28 | 70.5 | 13.0 |
| Ours | Mediastinum | 90 | 389 | 248 | 63.1 | 3.0 |
| Ours | Mediastinum | 90 | 389 | 305 | 78.0 | 6.0 |
| Ours | Abdomen | 86 | 595 | 419 | 70.1 | 5.1 |

**Optimal HOG Resolution:** The three slice-level classifiers $\mathcal{C}^1_{3\times3}$, $\mathcal{C}^1_{5\times5}$, and $\mathcal{C}^1_{9\times9}$ correspond to 9, 25 and 81-cell configurations of HOG respectively. **Fig. 4** shows both the training and six-fold CV performance for these three HOG configurations in the abdomen using mean-pooling. While the 81-cell configuration performs the best in training (using the highest resolution), it results in severe overfitting. The 25-cell con-figuration shows the best performance in the testing phase and is thus used in our final results.

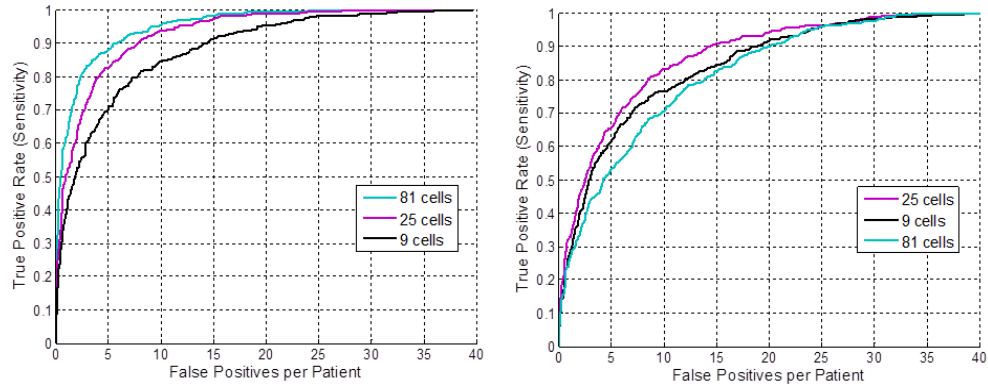

**Fig. 4.** Training (left) and six-fold cross-validation (right) FROC curves using the three HOG configurations for abdominal LN detection.

**Combining Datasets:** Medical imaging datasets are often much smaller than those used in natural image applications. In order to leverage the full capacity of our data, we train a model using both the abdominal and mediastinal LNs. Upon evaluation however, this combined model results in worse performance than the sectional models. Visualizations of both sectional models and their maximum response slices (**Fig. 5**) demonstrate the substantial difference in the average mediastinal and abdominal LN/background, leading to an ineffective combined model (when the linear classification contraint is imposed). Interestingly, the maximum response slice in the mediastinum is actually a FP (lung lesion). Our aggregation stage, a main contribution of this paper, helps overcome this relatively weak slice-level model.

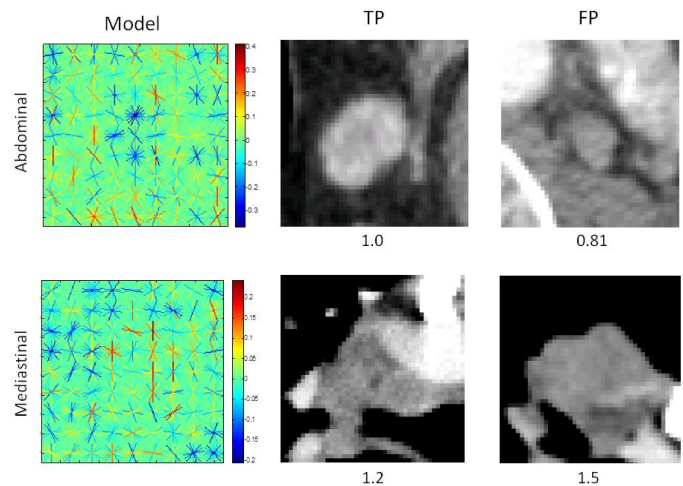

**Fig. 5.** Visualizations of the learned abdominal (top left) and mediastinal (bottom left) models with their maximum response TP (middle) and FP (right) slices and confidence scores.

## 4 Conclusion

We propose a novel approach to automated lymph node detection in CT images which exploits a hierarchy of classifiers trained on features extracted from 2D views of 3D candidate VOIs. In this manner, our detector circumvents expensive 3D feature computation during classification while still sufficiently capturing the spatial context necessary to recognize lymph node presence. Experimental results in both mediastinal and abdominal target regions demonstrate that our technique outperforms previous state-of-the-art methods for lymph node detection. A companion approach exploiting an alternative deep hierarchy for LN detection can be found in [16].

Ongoing work leverages an enhanced feature set derived from semantic contour cues [17, 18, 19], providing complementary mid-level information to the intensity gradients. Additionally, we construct a mixture of templates model by training LN size-specific classifiers. These methods allow for a substantial improvement in performance over the results reported here.

## Acknowledgements

This work was supported in part by the Intramural Research Program of the National Institutes of Health Clinical Center.